# Latent Tree Models and
# Approximate Inference in Bayesian Networks

**Yi Wang**                                                    WANGYI@CSE.UST.HK
**Nevin L. Zhang**                                             LZHANG@CSE.UST.HK
**Tao Chen**                                                   CSCT@CSE.UST.HK
*Department of Computer Science and Engineering*
*The Hong Kong University of Science and Technology*
*Clear Water Bay Road, Kowloon, Hong Kong, China*

## Abstract

We propose a novel method for approximate inference in Bayesian networks (BNs). The idea is to sample data from a BN, learn a latent tree model (LTM) from the data offline, and when online, make inference with the LTM instead of the original BN. Because LTMs are tree-structured, inference takes linear time. In the meantime, they can represent complex relationship among leaf nodes and hence the approximation accuracy is often good. Empirical evidence shows that our method can achieve good approximation accuracy at low online computational cost.

## 1. Introduction

*Latent tree models* (LTMs) are tree-structured Bayesian networks where leaf nodes represent *manifest variables* which are observed, while internal nodes represent *latent variables* which are hidden. They are previously known as hierarchical latent class models (Zhang, 2004). In this paper, we do not distinguish between variables and nodes, and assume that all variables are categorical.

Pearl (1988) was the first to identify LTMs as a potentially useful class of models. There are two reasons. First, inference in LTMs takes time linear in the number of nodes, while it is intractable in general BNs. Second, the latent variables capture complex relationships among the manifest variables. In an LTM, the manifest variables are mutually independent given the latent variables, while eliminating all the latent variables results in a completely connected BN.

We study the possibility of exploiting those two properties for approximate inference in BNs. Here is the most natural idea:

1. Offline: Obtain an LTM $\mathcal{M}$ that approximates a BN $\mathcal{N}$ in the sense that the joint distribution of the manifest variables in $\mathcal{M}$ approximately equals the joint distribution of the variables in $\mathcal{N}$.

2. Online: Use $\mathcal{M}$ instead of $\mathcal{N}$ to compute answers to probabilistic queries.

The cardinalities of the latent variables play a crucial role in the approximation scheme. They determine inferential complexity and influence approximation accuracy. At one extreme, we can represent a BN exactly using an LTM by setting the cardinalities of the latent variables large enough. In this case, the inferential complexity is very high. At the other extreme, we can set the cardinalities of the latent variables at 1. In this case, the manifest variables become mutually independent. The inferential complexity is the lowest and the approximation quality is the poorest. We seek an appropriate middle point between those two extremes.

We assume that there is a predetermined constraint on the cardinalities of the latent variables to control inferential complexity. We develop an algorithm for finding an LTM that satisfies the constraint and approximates the original BN well. The idea is to sample data from the BN, and learn an LTM from the data. The model structure is determined using hierarchical clustering of manifest variables. In each step, two closely correlated sets of manifest variables are grouped, and a new latent variable is introduced to account for the relationship between them. The cardinalities of the latent variables are set at the predetermined value. The parameters are optimized using the Expectation-Maximization (EM) algorithm (Dempster, Laird, & Rubin, 1977).

We have empirically evaluated our inference method on an array of networks. The possibility to tradeoff between inferential complexity and approximation accuracy has been demonstrated by adjusting the cardinality constraints. It turns out that our method is able to achieve good approximation accuracy before the cardinality becomes too high. We compared our method with loopy belief propagation (LBP) (Pearl, 1988), a standard approximate inference method which has been successfully used in many real world domains (Frey & MacKay, 1997; Murphy, Weiss, & Jordan, 1999). Given the same amount of time, our method achieves significantly higher accuracy than LBP in most cases. To achieve the same accuracy, LBP needs one to three orders of magnitude more time than our method.

Our inference method is fast because LTM is tree-structured. One can also construct a Chow-Liu tree (Chow & Liu, 1968) to approximate the original BN and use it for inference. We refer to this approach as the CL-based method. In comparison with our method, CL-based method is always faster, but it is not as accurate as our method.

Our scheme exploits the strong expressive capability of latent variable models. One can of course use other latent variable models instead of LTMs in the scheme. A straightforward choice is latent class model (LCM) (Hagenaars & McCutcheon, 2002). An LCM is an LTM with only one latent variable[1]. It assumes *local independence*, that is, the manifest variables are mutually independent conditioning on the latent variable. We also compare our method with this alternative. The results show that, under the same inferential complexity constraints, our method is more accurate than the LCM-based method.

It should be noted that our approximate scheme needs a lot of time in the offline phase. This is because that EM usually takes a long time to converge. Moreover, the time complexity of EM scales up linearly with the sample size, which should be set as large as possible to achieve high-quality approximation. Therefore, our method is suitable only for applications that allow a long offline phase.

The remainder of this paper is organized as follows. In Section 2, we review LTMs. In Section 3, we describe our method of constructing LTMs to approximate BNs. In Section 4, we describe our scheme for approximate inference formally. Section 5 reports empirical results. Section 6 discusses the relationship between our approach and existing work. Finally, in Section 7, we conclude this paper and point out some future directions.

---

1. In machine learning community, LCM is also referred to as naive Bayes model with latent variable.

## 2. Latent Tree Model

An LTM is a pair $\mathcal{M} = (m, \boldsymbol{\theta}_m)$. The first component $m$ denotes the rooted tree and the set of cardinalities of the latent variables. We will refer to $m$ as the model, and the rooted tree as the model structure. The second component $\boldsymbol{\theta}_m$ denotes the collection of parameters in $\mathcal{M}$. It contains a conditional probability table for each node given its parent.

Let $\mathbf{X}$ and $\mathbf{Y}$ be the set of manifest variables and the set of latent variables in $\mathcal{M}$, respectively. We use $P(\mathbf{X}, \mathbf{Y} | m, \boldsymbol{\theta}_m)$, or $P_{\mathcal{M}}(\mathbf{X}, \mathbf{Y})$ in short, to denote the joint distribution represented by $\mathcal{M}$. Two LTMs $\mathcal{M}$ and $\mathcal{M}'$ are *marginally equivalent* if they share the same set of manifest variables $\mathbf{X}$ and $P_{\mathcal{M}}(\mathbf{X}) = P_{\mathcal{M}'}(\mathbf{X})$. A model $m$ *includes* another model $m'$ if for any $\boldsymbol{\theta}_{m'}$ there exists $\boldsymbol{\theta}_m$ such that $(m, \boldsymbol{\theta}_m)$ and $(m', \boldsymbol{\theta}_{m'})$ are marginally equivalent. Two models $m$ and $m'$ are *marginally equivalent* if $m$ includes $m'$ and vice versa.

Let $|Z|$ denote the cardinality of a variable $Z$. For a node $Z$ in $m$, we use $\mathbf{nb}(Z)$ to denote the set of its neighbors. A model $m$ is *regular* if for any latent node $Y$,

1. If $Y$ has only two neighbors, then at least one of the neighbors must be a latent node and

$$|Y| < \frac{\prod_{Z \in \mathbf{nb}(Y)} |Z|}{\max_{Z \in \mathbf{nb}(Y)} |Z|}.$$

2. If $Y$ has more than two neighbors, then

$$|Y| \leq \frac{\prod_{Z \in \mathbf{nb}(Y)} |Z|}{\max_{Z \in \mathbf{nb}(Y)} |Z|}.$$

If a model $m$ is irregular, it is over-complicated. It can be reduced to a regular model $m'$ that is marginally equivalent to and contains fewer parameters than $m$ (Zhang, 2004). In a regular model, a latent node $Y$ is *saturated* if $|Y| = \frac{\prod_{Z \in \mathbf{nb}(Y)} |Z|}{\max_{Z \in \mathbf{nb}(Y)} |Z|}$. In this case, we say that $Y$ *subsumes* all its neighbors except the one with the largest cardinality.

## 3. Approximating Bayesian Networks with Latent Tree Models

In this section, we study the problem of approximating a BN with an LTM. Let $\mathcal{N}$ be the BN to be approximated. Let $\mathbf{X}$ be the set of variables in $\mathcal{N}$. For an LTM $\mathcal{M}$ to be an approximation of $\mathcal{N}$, it should use $\mathbf{X}$ as its manifest variables, and the cardinalities of its latent variables should not exceed a predetermined threshold $C$. Figure 1(b), 1(c), and 1(d) show three example LTMs that approximate the BN in Figure 1(a). They will be used to illustrate various steps in our method.

Let $P_{\mathcal{N}}(\mathbf{X})$ be the joint distribution represented by $\mathcal{N}$. An approximation $\mathcal{M}$ is of high quality if $P_{\mathcal{M}}(\mathbf{X})$ is close to $P_{\mathcal{N}}(\mathbf{X})$. We measure the quality of the approximation by the KL divergence (Cover & Thomas, 1991)

$$D[P_{\mathcal{N}}(\mathbf{X}) \| P_{\mathcal{M}}(\mathbf{X})] = \sum_{\mathbf{X}} P_{\mathcal{N}}(\mathbf{X}) \log \frac{P_{\mathcal{N}}(\mathbf{X})}{P_{\mathcal{M}}(\mathbf{X})}.$$

Our objective is to find an LTM that minimizes the KL divergence, i.e.,

$$\mathcal{M}^{\star} = \arg \min_{\mathcal{M}} D[P_{\mathcal{N}}(\mathbf{X}) \| P_{\mathcal{M}}(\mathbf{X})].$$

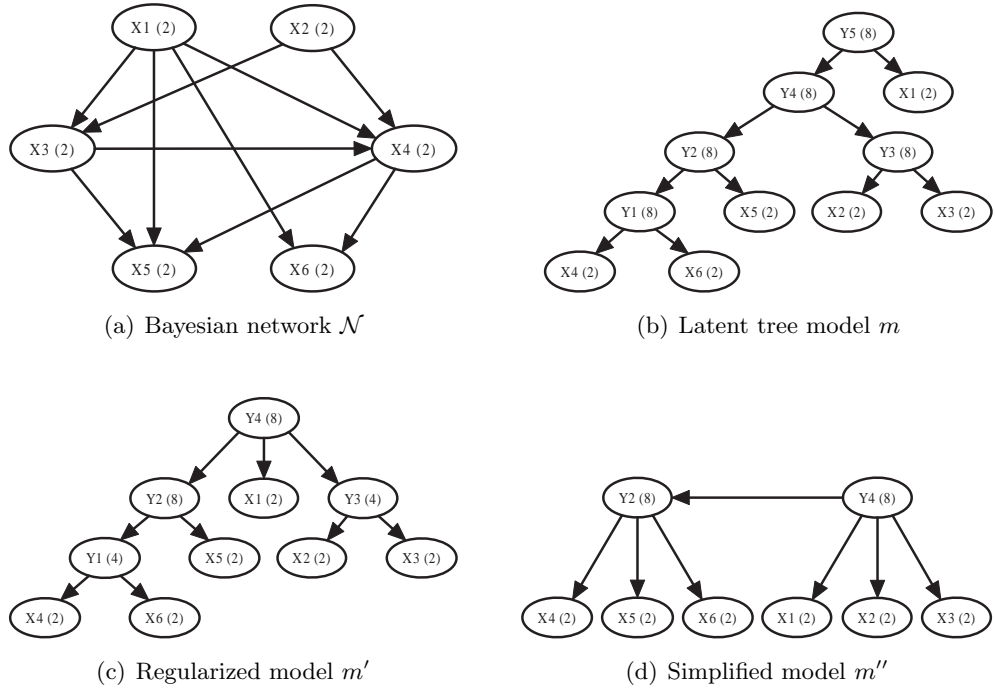

(a) Bayesian network $\mathcal{N}$

(b) Latent tree model $m$

(c) Regularized model $m'$

(d) Simplified model $m''$

Figure 1: An illustrative example. The numbers within the parentheses are the cardinalities of the variables.

An LTM $\mathcal{M}$ consists of two components, the model $m$ and the parameters $\boldsymbol{\theta}_m$. Therefore, the optimization problem can be naturally decomposed into two subproblems.

1. Find an optimal model $m^\star$.

2. Optimize the parameters $\boldsymbol{\theta}_m^\star$ for a given model $m$.

In the remainder of this section, we will discuss these two subproblems in details.

## 3.1 Parameter Optimization

We start by addressing the second subproblem. Given a model $m$, our target is to find

$$\boldsymbol{\theta}_m^\star = \arg\min_{\boldsymbol{\theta}_m} D[P_{\mathcal{N}}(\mathbf{X}) \| P(\mathbf{X}|m, \boldsymbol{\theta}_m)].$$

It turns out that, due to the presence of latent variables, the KL divergence is difficult to directly minimize. This can be seen by expanding the KL divergence as follows,

$$
\begin{aligned}
D[P_{\mathcal{N}}(\mathbf{X}) \| P(\mathbf{X}|m, \boldsymbol{\theta}_m)] &= \sum_{\mathbf{X}} P_{\mathcal{N}}(\mathbf{X}) \log \frac{P_{\mathcal{N}}(\mathbf{X})}{P(\mathbf{X}|m, \boldsymbol{\theta}_m)} \\
&= \sum_{\mathbf{X}} P_{\mathcal{N}}(\mathbf{X}) \log P_{\mathcal{N}}(\mathbf{X}) - \sum_{\mathbf{X}} P_{\mathcal{N}}(\mathbf{X}) \log P(\mathbf{X}|m, \boldsymbol{\theta}_m) \\
&= \sum_{\mathbf{X}} P_{\mathcal{N}}(\mathbf{X}) \log P_{\mathcal{N}}(\mathbf{X}) - \sum_{\mathbf{X}} P_{\mathcal{N}}(\mathbf{X}) \log \sum_{\mathbf{Y}} P(\mathbf{X}, \mathbf{Y}|m, \boldsymbol{\theta}_m).
\end{aligned}
$$

The first term on the last line can be neglected because it is independent of $\boldsymbol{\theta}_m$. The difficulty lies in maximizing the second term. The summation over latent variables $\mathbf{Y}$ appearing inside the logarithm makes this term indecomposable. Therefore, no closed-form solution can be obtained for $\boldsymbol{\theta}_m^\star$ by taking the derivative of this term with respect to $\boldsymbol{\theta}_m$ and setting it to zero.

We transform the problem into an asymptotically equivalent maximum likelihood estimation (MLE) problem. The idea is as follows.

1. Generate a data set $\mathcal{D}$ with $N$ independently and identically distributed samples from $P_{\mathcal{N}}(\mathbf{X})$.

2. Find the MLE of $\boldsymbol{\theta}_m$ with respect to $\mathcal{D}$, i.e.,

$$
\hat{\boldsymbol{\theta}}_m = \arg\max_{\boldsymbol{\theta}_m} P(\mathcal{D}|m, \boldsymbol{\theta}_m).
$$

It is well known that $\hat{\boldsymbol{\theta}}_m$ converges almost surely to $\boldsymbol{\theta}_m^\star$ as the sample size $N$ approaches infinity (Huber, 1967).

We now discuss the implementation of this solution. We start by generating $\mathcal{D}$ from $P_{\mathcal{N}}(\mathbf{X})$. Since $P_{\mathcal{N}}(\mathbf{X})$ is represented by BN $\mathcal{N}$, we use logic sampling (Henrion, 1988) for this task. Specifically, to generate a piece of sample from $P_{\mathcal{N}}(\mathbf{X})$, we process the nodes in a topological ordering[2]. When handling node $X$, we sample its value according to the conditional distribution $P(X|\pi(X) = j)$, where $\pi(X)$ denotes the set of parents of $X$ and $j$ denote their values that have been sampled earlier. To obtain $\mathcal{D}$, we repeat the procedure $N$ times.

Given $\mathcal{D}$, the next step is to find the MLE. Note that the values of latent variables $\mathbf{Y}$ are missing in $\mathcal{D}$. We thus use the EM algorithm (Dempster et al., 1977). Starting with a random guess, the EM algorithm iteratively improves the estimate until the change in loglikelihoods of two consecutive iterations is smaller than a predetermined threshold. A practical issue is that EM can converge to local maxima on the likelihood surface. The local maxima can be far from the global maxima, and thus can be poor approximations to $\boldsymbol{\theta}_m^\star$. Fortunately, the local maxima issue is not severe for LTMs (Wang & Zhang, 2006). In practice, one can also use various techniques such as multiple restart (Chickering & Heckerman, 1997) and data permutation (Elidan et al., 2002) to alleviate this issue.

Note that EM takes a long time to converge, especially when the sample size $N$ is large. This is why our algorithm has an expensive offline phase.

---

2. A topological ordering sorts the nodes in a DAG such that a node always precedes its children.

## 3.2 Exhaustive Search for the Optimal Model

We now consider the first subproblem, i.e., to find the best model $m^\star$. A straightforward way to solve this problem is to exhaust all possible models, find the optimal parameters $\boldsymbol{\theta}_m^\star$ for each model $m$, compute the KL divergence $D[P_\mathcal{N}(\mathbf{X}) \| P(\mathbf{X}|m, \boldsymbol{\theta}_m^\star)]$, and then return a model $m^\star$ with the minimum KL divergence.

The problem with this solution is its high computational complexity. Given a set of manifest variable $\mathbf{X}$, there are infinitely many models. One can always obtain new models by inserting latent variables to an existing model. As we will show in Section 3.5, it is sufficient to consider a finite subspace, i.e., the subspace of regular models. However, there are still super-exponentially many regular models (Zhang, 2004). For each model, we need to optimize its parameters by running EM, which is a time-consuming process. Therefore, the exhaustive search is computationally infeasible. In the following 4 subsections, we will present a heuristic method.

## 3.3 Heuristic Construction of Model Structure

We first present a heuristic for determining the model structure. In an LTM, two manifest variables are called siblings if they share the same parent. Our heuristic is based on two ideas: (1) In an LTM $\mathcal{M}$, siblings are generally more closely correlated than variables that are located far apart; (2) If $\mathcal{M}$ is a good approximation of $\mathcal{N}$, then two variables $X_i$ and $X_j$ are closely correlated in $\mathcal{M}$ if and only if they are closely correlated in $\mathcal{N}$. So we can examine each pair of variables in $\mathcal{N}$, pick the two variables that are most closely correlated, and introduce a latent variable as their parent in $\mathcal{M}$.

We measure the strength of correlation between a pair of variables $X_i$ and $X_j$ by the mutual information (Cover & Thomas, 1991)

$$I_\mathcal{N}(X_i; X_j) = \sum_{X_i, X_j} P_\mathcal{N}(X_i, X_j) \log \frac{P_\mathcal{N}(X_i, X_j)}{P_\mathcal{N}(X_i) P_\mathcal{N}(X_j)}.$$

To compute $I_\mathcal{N}(X_i; X_j)$, one need to make inference in $\mathcal{N}$. This could be computationally hard in the first place. So we use sampling technique to address this issue. Specifically, we generate a data set $\mathcal{D}$ with $N$ samples from the BN $\mathcal{N}$, and compute the empirical mutual information $\hat{I}(X_i; X_j)$ using the empirical distribution $\hat{P}(X_i, X_j)$ based on $\mathcal{D}$. By the strong law of large numbers, $\hat{I}(X_i; X_j)$ will almost surely converge to $I_\mathcal{N}(X_i; X_j)$ as the sample size $N$ goes to infinity.

We now use the BN shown in Figure 1(a) as an example to illustrate the idea. It contains 6 binary variables $X_1$, $X_2$, ..., $X_6$. Suppose the empirical mutual information based on some data set $\mathcal{D}$ is as presented in Table 1. As discussed above, we regard those as approximation to mutual information between variables in $\mathcal{N}$ and hence regard them as approximation to mutual information between variables in the final LTM $\mathcal{M}$ that we are to construct. We find that $X_4$ and $X_6$ are the pair with the largest mutual information. Therefore, we create a latent variable $Y_1$ and make it the parent of $X_4$ and $X_6$.

The next step is to find, among $Y_1$, $X_1$, $X_2$, $X_3$, and $X_5$, the pair of variables with the largest mutual information in $\mathcal{M}$. There is one difficulty: $Y_1$ is not in the original Bayesian network and hence not observed in the data set. The mutual information between $Y_1$ and the other variables cannot be computed directly. We hence seek an approximation. In the final model $\mathcal{M}$, $Y_1$ would d-separate $X_4$ and $X_6$ from the other variables. Therefore, for any $X \in \{X_1, X_2, X_3, X_5\}$, we have

$$I_{\mathcal{M}}(Y_1; X) \geq I_{\mathcal{M}}(X_4; X), \ \ I_{\mathcal{M}}(Y_1; X) \geq I_{\mathcal{M}}(X_6; X).$$

|       | $X_1$  | $X_2$  | $X_3$  | $X_4$  | $X_5$  |
|-------|--------|--------|--------|--------|--------|
| $X_1$ | -      | -      | -      | -      | -      |
| $X_2$ | 0.0000 | -      | -      | -      | -      |
| $X_3$ | 0.0003 | 0.0971 | -      | -      | -      |
| $X_4$ | 0.0015 | 0.0654 | 0.0196 | -      | -      |
| $X_5$ | 0.0017 | 0.0311 | 0.0086 | 0.1264 | -      |
| $X_6$ | 0.0102 | 0.0252 | 0.0080 | 0.1817 | 0.0486 |

Table 1: Empirical mutual information between manifest variables

We hence approximate $I_{\mathcal{M}}(Y_1; X)$ using the lower bound

$$\max\{I_{\mathcal{M}}(X_4; X), I_{\mathcal{M}}(X_6; X)\}.$$

Back to our running example, the estimated mutual information between $Y_1$ and $X_1$, $X_2$, $X_3$, $X_5$ is as presented in Table 2. We see that the next pair to pick is $Y_1$ and $X_5$. We introduce a latent variable $Y_2$ as the parent of $Y_1$ and $X_5$. The process continues. The final model structure is a binary tree as shown in Figure 1(b).

|       | $X_1$  | $X_2$  | $X_3$  | $X_5$  |
|-------|--------|--------|--------|--------|
| $Y_1$ | 0.0102 | 0.0654 | 0.0196 | 0.1264 |

Table 2: Estimated mutual information between $Y_1$ and manifest variables

## 3.4 Cardinalities of Latent Variables

After obtaining a model structure, the next step is to determine the cardinalities of the latent variables. We set the cardinalities of all the latent variables at a predetermined value $C$. In the following, we discuss how the choice of $C$ influences quality of approximation and inferential efficiency.

We first discuss the impact of the value of $C$ on the approximation quality. We start by considering the case when $C$ equals to $C_{\max} = \prod_{X \in \mathbf{X}} |X|$, i.e., the product of the cardinalities of all the manifest variables. In this case, each latent variable can be viewed as a joint variable of all the manifest variables. We can therefore set the parameters $\boldsymbol{\theta}_m$ so that $P(\mathbf{X}|m, \boldsymbol{\theta}_m) = P_{\mathcal{N}}(\mathbf{X})$. That is, $m$ can capture all the interactions among the manifest variables.

What happens if we decrease $C$? It can be shown that the approximation quality will degrade. Let $m$ be a model obtained with value $C$ and $m'$ be another model obtained with a smaller value $C'$. It is easy to see that $m$ includes $m'$. The following lemma states that the approximation quality of $m'$ is no better than that of $m$.

**Lemma 1** *Let $P(\mathbf{X})$ be a joint probability distribution of $\mathbf{X}$. Let $m$ and $m'$ be two models with manifest variables $\mathbf{X}$. If $m$ includes $m'$, then*

$$\min_{\boldsymbol{\theta}_m} D[P(\mathbf{X}) \| P(\mathbf{X}|m, \boldsymbol{\theta}_m)] \leq \min_{\boldsymbol{\theta}_{m'}} D[P(\mathbf{X}) \| P(\mathbf{X}|m', \boldsymbol{\theta}_{m'})].$$

**Proof:** Define

$$\boldsymbol{\theta}_{m'}^\star = \arg\min_{\boldsymbol{\theta}_{m'}} D[P(\mathbf{X}) \| P(\mathbf{X}|m', \boldsymbol{\theta}_{m'})].$$

Because $m$ includes $m'$, there must be parameters $\boldsymbol{\theta}_m^\star$ of $m$ such that

$$P(\mathbf{X}|m, \boldsymbol{\theta}_m^\star) = P(\mathbf{X}|m', \boldsymbol{\theta}_{m'}^\star).$$

Therefore,

$$
\begin{aligned}
\min_{\boldsymbol{\theta}_m} D[P(\mathbf{X}) \| P(\mathbf{X}|m, \boldsymbol{\theta}_m)] &\leq D[P(\mathbf{X}) \| P(\mathbf{X}|m, \boldsymbol{\theta}_m^\star)] \\
&= D[P(\mathbf{X}) \| P(\mathbf{X}|m', \boldsymbol{\theta}_{m'}^\star)] \\
&= \min_{\boldsymbol{\theta}_{m'}} D[P(\mathbf{X}) \| P(\mathbf{X}|m', \boldsymbol{\theta}_{m'})]
\end{aligned}
$$

**Q.E.D.**

As mentioned earlier, when $C$ is large enough, model $m$ can capture all the interactions among the manifest variables and hence can represent the joint distribution $P_{\mathcal{N}}(\mathbf{X})$ exactly. If $C$ is not large enough, we can only represent $P_{\mathcal{N}}(\mathbf{X})$ approximately. According to the previous discussion, as $C$ decreases, the approximation accuracy (in terms of KL divergence) will gradually degrade, indicating that model $m$ can capture less and less interactions among the manifest variables. The worst case occurs when $C = 1$. In this case, all the interactions are lost. The approximation accuracy is the poorest.

The parameter $C$ also determines the computational cost of making inference in $m$. We use the clique tree propagation (CTP) algorithm for inference. So we measure the cost by the *inferential complexity*, which is defined to be the sum of the clique sizes in the clique tree of $m$. It is given by

$$(|\mathbf{X}| - 2) \cdot C^2 + \sum_{X \in \mathbf{X}} |X| \cdot C. \tag{1}$$

Note that $|\mathbf{X}|$ is the number of manifest variables, while $|X|$ is the cardinality of a manifest variable $X$. Therefore, one can control the inferential complexity by changing the value of $C$. The smaller the value of $C$, the lower the complexity.

In summary, one can achieve a tradeoff between the approximation quality and the inferential complexity of the resultant model $m$ by tuning the parameter $C$. In Figure 1(b), we set $C = 8$.

## 3.5 Model Regularization

Suppose we have obtained a model $m$ using the technique described in Section 3.3 and by setting the cardinalities of the latent variables at a certain value. In the following two subsections, we will show that it is sometimes possible to simplify $m$ without compromising the approximation quality.

We first notice that $m$ could be irregular. As an example, let us consider the model in Figure 1(b). It is constructed as an approximation to the BN $\mathcal{N}$ in Figure 1(a) with $C = 8$. By checking the latent variables, we find that $Y_5$ violates the regularity condition. It has only two neighbors and $|Y_5| \geq |X_1| \cdot |Y_4| / \max\{|X_1|, |Y_4|\}$. $Y_1$ and $Y_3$ also violate the regularity condition because $|Y_1| > |X_4| \cdot |X_6| \cdot |Y_2| / \max\{|X_4|, |X_6|, |Y_2|\}$ and $|Y_3| > |X_2| \cdot |X_3| \cdot |Y_4| / \max\{|X_2|, |X_3|, |Y_4|\}$. The following proposition suggests that irregular models should always be simplified until they become regular.

**Proposition 1** *If $m$ is an irregular model, then there must exists a model $m'$ with lower inferential complexity such that*

$$\min_{\boldsymbol{\theta}_m} D[P_\mathcal{N}(\mathbf{X}) \| P(\mathbf{X}|m, \boldsymbol{\theta}_m)] = \min_{\boldsymbol{\theta}_{m'}} D[P_\mathcal{N}(\mathbf{X}) \| P(\mathbf{X}|m', \boldsymbol{\theta}_{m'})]. \tag{2}$$

**Proof:** Let $Y$ be a latent variable in $m$ which violates the regularity condition. Denote its neighbors by $Z_1, Z_2, \ldots, Z_k$. We define another model $m'$ as follows:

1. If $Y$ has only two neighbors, then remove $Y$ from $m$ and connect $Z_1$ with $Z_2$.

2. Otherwise, replace $Y$ with a saturated latent variable $Y'$, i.e.,

$$|Y'| = \frac{\prod_{i=1}^k |Z_i|}{\max_{i=1}^k |Z_i|}.$$

As shown by Zhang (2004), for any parameters $\boldsymbol{\theta}_m$ of $m$, there exists parameters $\boldsymbol{\theta}_{m'}$ of $m'$ such that $(m, \boldsymbol{\theta}_m)$ and $(m', \boldsymbol{\theta}_{m'})$ are marginally equivalent. The reverse is also true. Therefore, $m$ and $m'$ are marginally equivalent. Equation 2 thus follows from Lemma 1.

To show that the inferential complexity of $m'$ is lower than that of $m$, we compare the clique trees of $m$ and $m'$. Consider the aforementioned two cases:

1. $Y$ has only two neighbors. In this case, cliques $\{Y, Z_1\}$ and $\{Y, Z_2\}$ in the clique tree of $m$ are replaced with $\{Z_1, Z_2\}$ in the clique tree of $m'$. Assume $|Z_2| \geq |Z_1|$. The difference in the sum of clique sizes is

$$
\begin{aligned}
sum(m) - sum(m') &= |Y||Z_1| + |Y||Z_2| - |Z_1||Z_2| \\
&\geq |Z_1||Z_1| + |Z_1||Z_2| - |Z_1||Z_2| \\
&= |Z_1||Z_1| \\
&> 0.
\end{aligned}
$$

2. $Y$ has more than two neighbors. In this case, for all $i = 1, 2, \ldots, k$, clique $\{Y, Z_i\}$ in the clique tree of $m$ is replaced with a smaller clique $\{Y', Z_i\}$ in the clique tree of $m'$.

In both cases, the inferential complexity of $m'$ is lower than that of $m$.
**Q.E.D.**

The proof of Proposition 1 presents a way to handle a latent variable that violates the regularity condition, i.e., either eliminating it or decreasing its cardinality. To regularize an irregular model, we handle all the latent variables in the order by which they are created in Section 3.3. In the following, we use the irregular model $m$ in Figure 1(b) to demonstrate the regularization process.

We begin with latent variable $Y_1$. It has three neighbors and violates the regularity condition. So we decrease its cardinality to $|X_4| \cdot |X_6| \cdot |Y_2| / \max\{|X_4|, |X_6|, |Y_2|\} = 4$. Then we consider $Y_2$. It satisfies the regularity condition and hence no changes are made. The next latent variable to examine is $Y_3$. It violates the regularity condition. So we decrease its cardinality to $|X_2| \cdot |X_3| \cdot |Y_4| / \max\{|X_2|, |X_3|, |Y_4|\} = 4$. We do not change $Y_4$ because it does not cause irregularity. At last, we remove $Y_5$, which has only two neighbors and violates the regularity condition, and connect $Y_4$ with $X_1$. We end up with the regular model $m'$ as shown in Figure 1(c).

## 3.6 Further Simplifications

After regularization, there are sometimes still opportunities for further model simplification. Take the model $m'$ in Figure 1(c) as an example. It contains two adjacent latent variables $Y_1$ and $Y_2$. Both variables are saturated. $Y_1$ subsumes $X_4$ and $X_6$, and $Y_2$ subsumes $Y_1$ and $X_5$. $Y_2$ can be viewed as a joint variable of $Y_1$ and $X_5$, while $Y_1$ can be in turn viewed as a joint variable of $X_4$ and $X_6$. Intuitively, we can eliminate $Y_1$ and directly make $Y_2$ the joint variable of $X_4$, $X_5$, and $X_6$. This intuition is formalized by the following proposition.

**Proposition 2** *Let $m$ be a model with more than one latent node. Let $Y_1$ and $Y_2$ be two adjacent latent nodes. If both $Y_1$ and $Y_2$ are saturated while $Y_2$ subsumes $Y_1$, then there exist another model $m'$ that is marginally equivalent to and has lower inferential complexity than $m$. Therefore,*

$$\min_{\boldsymbol{\theta}_m} D[P_{\mathcal{N}}(\mathbf{X}) \| P(\mathbf{X}|m, \boldsymbol{\theta}_m)] = \min_{\boldsymbol{\theta}_{m'}} D[P_{\mathcal{N}}(\mathbf{X}) \| P(\mathbf{X}|m', \boldsymbol{\theta}_{m'})].$$

**Proof:** We enumerate the neighbors of $Y_1$ as $Y_2, Z_{11}, Z_{12}, \ldots, Z_{1k}$, and the neighbors of $Y_2$ as $Y_1, Z_{21}, Z_{22}, \ldots, Z_{2l}$. Define another model $m'$ by removing $Y_1$ from $m$ and connecting $Z_{11}, Z_{12}, \ldots, Z_{1k}$ to $Y_2$. See Figure 2. We now prove that $m$ and $m'$ are marginally equivalent, while the inferential complexity of $m'$ is lower than that of $m$.

We start by proving the marginal equivalence. For technical convenience, we will work with unrooted models. An unrooted model is a model with all directions on the edges dropped. Parameters of an unrooted model include a potential for each edge in the model. The potential is a non-negative function of the two variables that are connected by the edge. The concept of marginal equivalence can be defined the same way as for rooted models.

As shown by Zhang (2004), a model is marginally equivalent to its unrooted version. Therefore, to prove the marginal equivalence between $m$ and $m'$, it is sufficient to show that the unrooted versions of $m$ and $m'$ are marginally equivalent. For simplicity, we abuse $m$ and $m'$ to denote the unrooted models. We also use $f(\cdot)$ to denote a potential in $\boldsymbol{\theta}_m$, and $g(\cdot)$ to denote a potential in $\boldsymbol{\theta}_{m'}$.

Note that $Y_1$ and $Y_2$ are saturated, while $Y_2$ subsumes $Y_1$. When all variables have no less than two states, this implies that:

1. $Y_1$ subsumes $Z_{11}, Z_{12}, \ldots, Z_{1k}$.

2. Suppose that $|Z_{2l}| = \max_{j=1}^{l} |Z_{2j}|$. Then $Y_2$ subsumes $Z_{21}, Z_{22}, \ldots, Z_{2l-1}$.

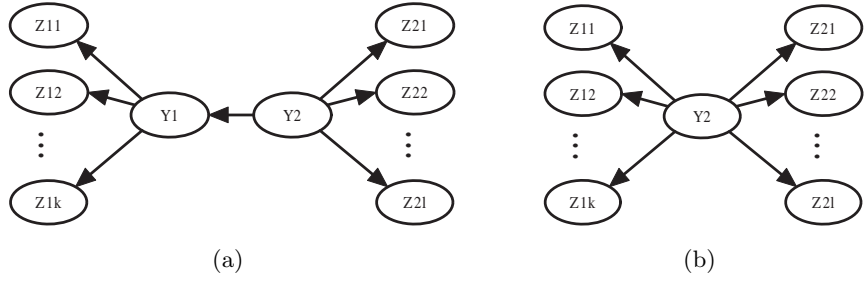

(a)                              (b)

Figure 2: Further simplification. (a) A part of a model that contains two adjacent and saturated latent nodes $Y_1$ and $Y_2$, with $Y_2$ subsuming $Y_1$. (b) Simplified model with $Y_1$ eliminated.

Therefore, a state of $Y_1$ can be written as $y_1 = <z_{11}, z_{12}, \ldots, z_{1k}>$, while a state of $Y_2$ can be written as $y_2 = <y_1, z_{21}, z_{22}, \ldots, z_{2l-1}>$. The latter can be further expanded as $y_2 = <z_{11}, z_{12}, \ldots, z_{1k}, z_{21}, z_{22}, \ldots, z_{2l-1}>$.

We first show that $m'$ includes $m$. Let $\boldsymbol{\theta}_m$ be parameters of $m$. We define parameters $\boldsymbol{\theta}_{m'}$ of $m'$ as follows:

- Potential for edge $Y_2 — Z_{2l}$:

$$g(Y_2 = <z_{11}, z_{12}, \ldots, z_{1k}, z_{21}, z_{22}, \ldots, z_{2l-1}>, Z_{2l} = z_{2l})$$
$$= \sum_{Y_1, Y_2} f(Y_1, Y_2) \prod_{i=1}^{k} f(Y_1, Z_{1i} = z_{1i}) \prod_{j=1}^{l} f(Y_2, Z_{2j} = z_{2j}).$$

- Potential for edge $Y_2 — Z_{1i}, \forall i = 1, 2, \ldots, k$:

$$g(Y_2 = <z_{11}, z_{12}, \ldots, z_{1k}, z_{21}, z_{22}, \ldots, z_{2l-1}>, Z_{1i} = z'_{1i}) = \begin{cases} 1 & z_{1i} = z'_{1i} \\ 0 & \text{otherwise} \end{cases}$$

- Potential for edge $Y_2 — Z_{2j}, \forall j = 1, 2, \ldots, l-1$:

$$g(Y_2 = <z_{11}, z_{12}, \ldots, z_{1k}, z_{21}, z_{22}, \ldots, z_{2l-1}>, Z_{2j} = z'_{2j}) = \begin{cases} 1 & z_{2j} = z'_{2j} \\ 0 & \text{otherwise} \end{cases}$$

- Set the other potentials in $\boldsymbol{\theta}_{m'}$ the same as those in $\boldsymbol{\theta}_m$.

It is easy to verify that

$$\sum_{Y_1, Y_2} f(Y_1, Y_2) \prod_{i=1}^{k} f(Y_1, Z_{1i}) \prod_{j=1}^{l} f(Y_2, Z_{2j}) = \sum_{Y_2} \prod_{i=1}^{k} g(Y_2, Z_{1i}) \prod_{j=1}^{l} g(Y_2, Z_{2j}). \tag{3}$$

Therefore,

$$P(\mathbf{X}|m, \boldsymbol{\theta}_m) = P(\mathbf{X}|m', \boldsymbol{\theta}_{m'}). \tag{4}$$

Next, we prove that $m$ includes $m'$. Given parameters $\boldsymbol{\theta}_{m'}$ of $m'$, we define parameters $\boldsymbol{\theta}_m$ of $m$ as follows:

- Potential for edge $Y_1 — Y_2$:

$$f(Y_1 =< z_{11}, z_{12}, \ldots, z_{1k} >, Y_2 = y_2) = \prod_{i=1}^{k} g(Y_2 = y_2, Z_{1i} = z_{1i}).$$

- Potential for edge $Y_1 — Z_{1i}, \forall i = 1, 2, \ldots, k$:

$$f(Y_1 =< z_{11}, z_{12}, \ldots, z_{1k} >, Z_{1i} = z'_{1i}) = \left\{ \begin{array}{ll} 1 & z_{1i} = z'_{1i} \\ 0 & \text{otherwise} \end{array} \right.$$

- Set the other potentials in $\boldsymbol{\theta}_m$ the same as those in $\boldsymbol{\theta}_{m'}$.

It can be verified that Equation 3 and 4 also hold. Therefore, $m$ and $m'$ are marginally equivalent.

We now compare the inferential complexity of $m$ and $m'$. According to the construction of $m'$, the clique tree of $m'$ is different from the clique tree of $m$ in that it contains one less clique $\{Y_1, Y_2\}$ and replaces clique $\{Y_1, Z_{1i}\}$ with $\{Y_2, Z_{1i}\}$ for all $i = 1, 2, \ldots, k$. Therefore, the difference between the the sum of clique sizes is

$$
\begin{array}{rcl}
sum(m) - sum(m') & = & |Y_1||Y_2| + \sum_i |Y_1||Z_{1i}| - \sum_i |Y_2||Z_{1i}| \\
& = & |Y_2| \prod_i |Z_{1i}| + \sum_i |Y_1||Z_{1i}| - \sum_i |Y_2||Z_{1i}| \\
& = & |Y_2|(\prod_i |Z_{1i}| - \sum_i |Z_{1i}|) + \sum_i |Y_1||Z_{1i}|.
\end{array}
$$

The first term on the last line is non-negative because $\prod_i |Z_{1i}| \geq \sum_i |Z_{1i}|$ when $|Z_{1i}| \geq 2$ for all $i = 1, 2, \ldots, k$. Therefore, the inferential complexity of $m'$ is always lower than that of $m$ when $Z_{1i}$ is nontrivial.
**Q.E.D.**

Given the regularized model, we check each pair of adjacent latent variables and apply Proposition 2 to eliminate redundant latent variables. We use the model $m'$ in Figure 1(c) as an example to demonstrate the process. The first pair to check are $Y_1$ and $Y_2$. Both of them are saturated while $Y_2$ subsumes $Y_1$. We thus remove $Y_1$ and connect $Y_2$ to $X_4$ and $X_6$. We then check $Y_3$ and $Y_4$. It turns out that $Y_3$ is redundant. Therefore, we remove it and connect $Y_4$ to $X_2$ and $X_3$. The last pair to check are $Y_2$ and $Y_4$. They are both saturated, but neither of them subsumes the other. Hence, they cannot be removed. The final model $m''$ is shown in Figure 1(d).

### 3.7 The Algorithm LTAB

To summarize, we have outlined an algorithm for approximating BNs using LTMs. We call the algorithm LTAB, a shorthand for **L**atent **T**ree **A**pproximation of **B**ayesian network. It has 3 inputs: a BN $\mathcal{N}$, a predetermined cardinality $C$ for latent variables, and a sample size $N$. The output of `LTAB` is an LTM that approximates $P_\mathcal{N}(\mathbf{X})$, the joint probability distribution represented by $\mathcal{N}$. `LTAB` is briefly described as follows.

1. Generate a data set $\mathcal{D}$ of $N$ i.i.d. samples from $P_\mathcal{N}(\mathbf{X})$. (Section 3.1)

2. Obtain an LTM structure by performing hierarchical clustering of variables, using empirical mutual information based on $\mathcal{D}$ as the similarity measure. (Section 3.3)

3. Set cardinalities of latent variables at $C$ and simplify the model. (Section 3.4 – 3.6)

4. Optimize parameters by running EM. (Section 3.1)

5. Return the resultant LTM.

## 4. LTM-based Approximate Inference

The focus of this paper is approximate inference in Bayesian networks. We propose the following two-phase method:

1. Offline: Given a BN $\mathcal{N}$, use `LTAB` to construct an approximation $\mathcal{M}$. The sample size $N$ should be set as large as possible, while the cardinality $C$ should be determined to meet the requirement on inferential complexity.

2. Online: Make inference in $\mathcal{M}$ instead of $\mathcal{N}$. More specifically, given a piece of evidence $\mathbf{E} = \mathbf{e}$ and a querying variable $Q$, return $P_\mathcal{M}(Q|\mathbf{E} = \mathbf{e})$ as an approximation to $P_\mathcal{N}(Q|\mathbf{E} = \mathbf{e})$.

## 5. Empirical Results

In this section, we empirically evaluate our approximate inference method. We first examine the impact of sample size $N$ and cardinality $C$ on the performance of our method. Then we compare the our method with CTP, LBP, the CL-based method, and the LCM-based method.

We used 8 networks in our experiments. They are listed in Table 3. `CPCS54` is a subset of the `CPCS` network (Pradhan et al., 1994). The other networks are available at `http://www.cs.huji.ac.il/labs/compbio/Repository/`. Table 3 also reports the characteristics of the networks, including the number of nodes, the average/max indegree and cardinality of the nodes, and the inferential complexity (i.e., the sum of the clique sizes in the clique tree). The networks are sorted in ascending order with respect to the inferential complexity.

For each network, we simulated 500 pieces of evidence. Each piece of evidence was set on all the leaf nodes by sampling based on the joint probability distribution. Then we used the CTP algorithm and the approximate inference methods to compute the posterior distribution of each non-leaf node conditioned on each piece of evidence. The accuracy of an approximate method is measured by the average KL divergence between the exact and the approximate posterior distributions over all the query nodes and evidence.

All the algorithms in the experiments were implemented in Java and run on a machine with an Intel Pentium IV 3.2GHz CPU and 1GB RAM.

| Network | Number of Nodes | Average/Max Indegree | Average/Max Cardinality | Inferential Complexity |
|---------|-----------------|----------------------|-------------------------|------------------------|
| ALARM | 37 | 1.24/4 | 2.84/4 | 1,038 |
| WIN95PTS | 76 | 1.47/7 | 2/2 | 2,684 |
| HAILFINDER | 56 | 1.18/4 | 3.98/11 | 9,706 |
| INSURANCE | 27 | 1.93/3 | 3.3/5 | 29,352 |
| CPCS54 | 54 | 2/9 | 2/2 | 109,208 |
| WATER | 32 | 2.06/5 | 3.62/4 | 3,028,305 |
| MILDEW | 35 | 1.31/3 | 17.6/100 | 3,400,464 |
| BARLEY | 48 | 1.75/4 | 8.77/67 | 17,140,796 |

Table 3: Networks and their characteristics.

## 5.1 Impact of $N$ and $C$

We discussed the impact of $N$ and $C$ on the performance of our method in Section 3. This subsection empirically verifies the claims.

Three sample sizes were chosen in the experiments: $1k$, $10k$, and $100k$. For each network, we also chose a set of $C$. LTMs were then learned using LTAB with different combination of the values of $N$ and $C$. For parameter learning, we terminated EM either when the improvement in loglikelihoods is smaller than 0.1, or when the algorithm ran for two months. The multiple restarting strategy by Chickering and Heckerman (1997) was used to avoid local maxima. The number of starting points was set at 16.

The running time of LTAB is plotted in Figure 3. The $y$-axes denote the time in hours, while the $x$-axes denote the parameter $C$ for LTAB. The three curves correspond to different values of $N$. In general, the running time increases with $N$ and $C$, ranging from seconds to weeks. For some settings, EM failed to converge in two months. Those settings are indicated by arrows in the plots. We emphasize that LTAB is executed offline and its running time should not be confused with the time for online inference, which will be reported next.

After obtaining the LTMs, we used clique tree propagation to make inference. The approximation accuracy are shown in Figure 4. The $y$-axes denote the average KL divergence, while the $x$-axes still denote the parameter $C$ for LTAB. There are five curves and one horizontal line in each plot. The three curves labeled as LTM are for our method, which correspond to the three sample sizes we used. The remaining two curves and the horizontal line are for the other approximate inference methods. We will discuss them in Sections 5.3 – 5.5.

We first examine the impact of sample size by comparing the corresponding curves in each plot. We find that, in general, the curves for larger samples are located below those for smaller ones. This shows that the approximation accuracy increases with the size of the training data.

To see the impact of $C$, we examine each individual curve from left to right. According to our discussion, the curve is expected to drop monotonically as $C$ increases. This is generally true for the results with sample size $100k$. For sample sizes $1k$ and $10k$, however, there are cases in which the approximation becomes poorer as $C$ increases. See Figure 4(e) and 4(f). This phenomenon does not conflict with our claims. As $C$ increases, the expressive power of the learned LTM increases. So it tends to overfit the data. On the other hand, the empirical distribution of a small set of data may significantly deviate from the joint distribution of the BN. This also suggests that the sample size should be set as large as possible.

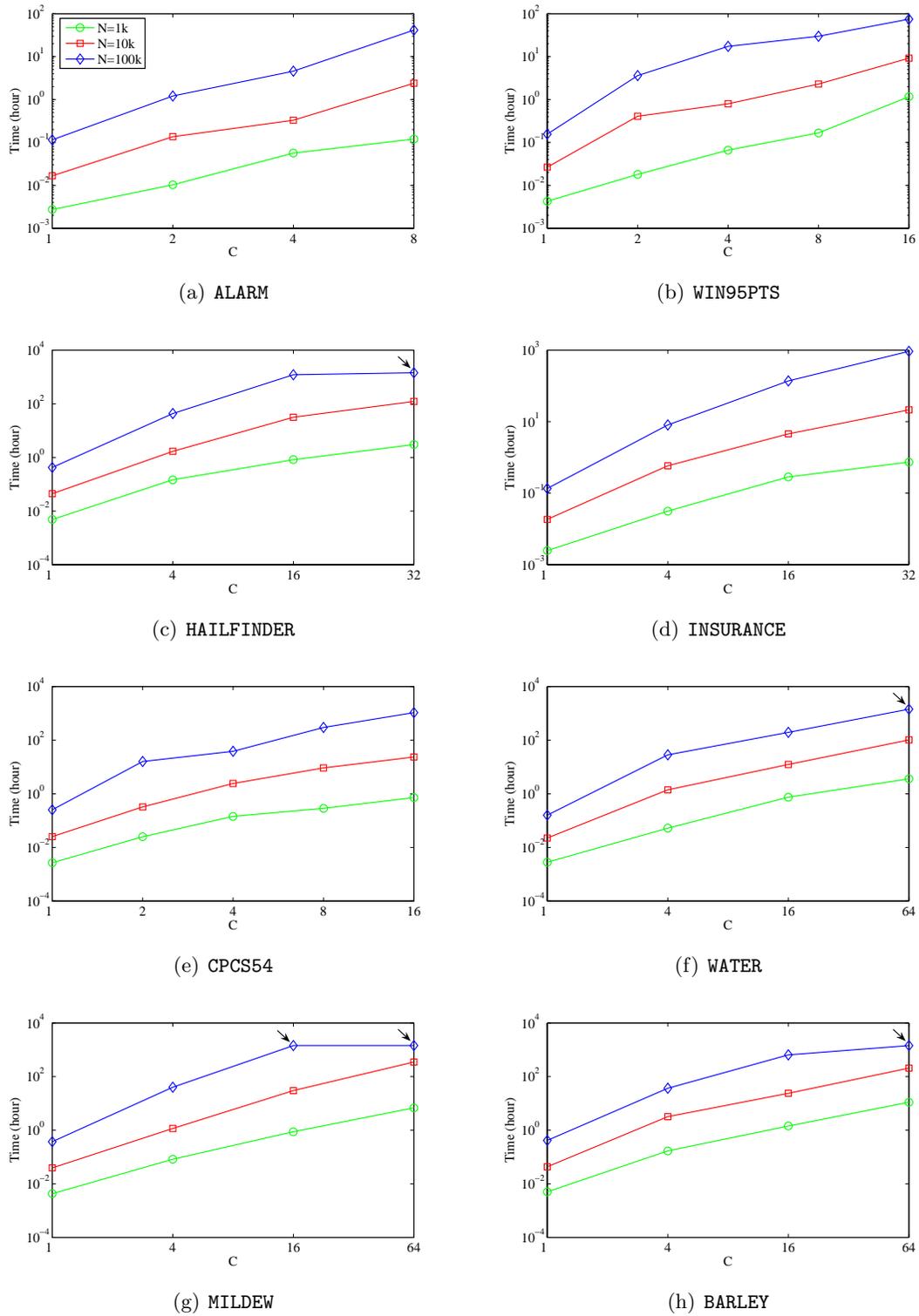

Figure 3: Running time of LTAB. Settings for which EM did not converge are indicated by arrows.

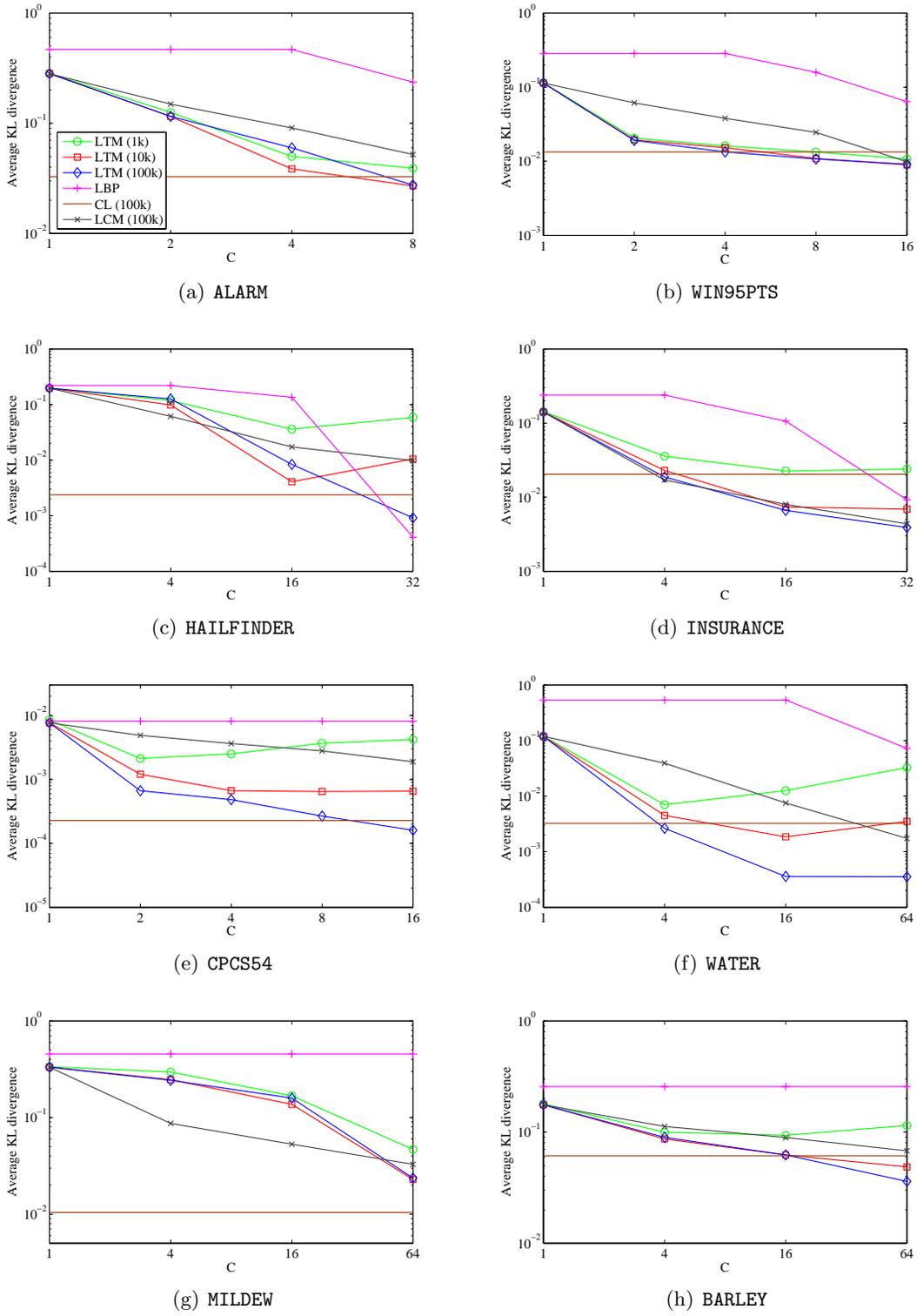

Figure 4: Approximation accuracy.

Finally, let us examine the impact of $N$ and $C$ on the inferential complexity. Figure 5 plots the running time for different methods to answer all the queries. For now, we only consider the three curves that are labeled as LTM. It can be seen that the three curves overlap in all plots. This implies that the running time is independent of the sample size $N$. On the other hand, all the curves are monotonically increasing. This confirms our claim that the inferential complexity is positively dependent on $C$.

In the following subsections, if not stated explicitly otherwise, we will only consider the results for $N = 100k$ and the largest $C$. Under these settings, our method achieves the highest accuracy.

## 5.2 Comparison with CTP

We now compare our method with CTP, a state-of-the-art exact inference algorithm. The first concern is that, how accurate is our method. By examining Figure 4, we argue that our method always achieves good approximation accuracy: For HAILFINDER, CPCS54, WATER, the average KL divergence of our method is around or less than $10^{-3}$; For the other networks, the average KL divergence is around or less than $10^{-2}$.

We next compare the inferential efficiency of our method and the CTP algorithm. The running time of CTP is denoted by dashed horizontal lines in the plots of Figure 5. It can be seen that our method is more efficient than the CTP algorithm. In particular, for the five networks with the highest inferential complexity, our method is faster than CTP by two to three orders of magnitude.

To summarize, the results suggest that our method can achieve good approximation accuracy at low computational cost.

## 5.3 Comparison with LBP

We now compare our method with LBP. The latter is an iterative algorithm. It can be used as an anytime inference method by running a specific number of iterations. In our first set of experiments, we let LBP run as long as our method and compare their approximation accuracy. We did this for each network and each value of $C$. The accuracy of LBP are denoted by the curves labeled as LBP in Figure 4. By comparing those curves with the LTM curves for $N = 100k$, we see that our method achieves significantly higher accuracy than LBP in most cases: For WATER, the difference in average KL divergence is up to three orders of magnitude; For the other networks, the difference is up to one order of magnitude. For HAILFINDER with $C = 32$, LBP is two times more accurate than our method. However, our method also achieves good approximation accuracy in this case. The average KL divergence is smaller than $10^{-3}$. Finally, we noticed that LBP curves are horizontal lines for CPCS54, MILDEW, and BARLEY. Further investigation on those cases shows that LBP finished only one iteration in the given time period.

We next examine how much time it takes for LBP to achieve the same level of accuracy as our method. For each piece of evidence, we ran LBP until its average KL divergence is comparable with that of our method or the number of iterations exceeds 100. The running time of LBP are denoted by the curves labeled as LBP in Figure 5. Comparing those curves with the LTM curves, we found that LBP takes much more time than our method: For MILDEW, LBP is slower than our method by three orders of magnitude; For the other networks except HAILFINDER, LBP is slower by one to two orders of magnitude; For HAILFINDER with $C = 32$, the running time of the two methods are similar. The results show that our method compares more favorably to LBP in the networks that we examined.

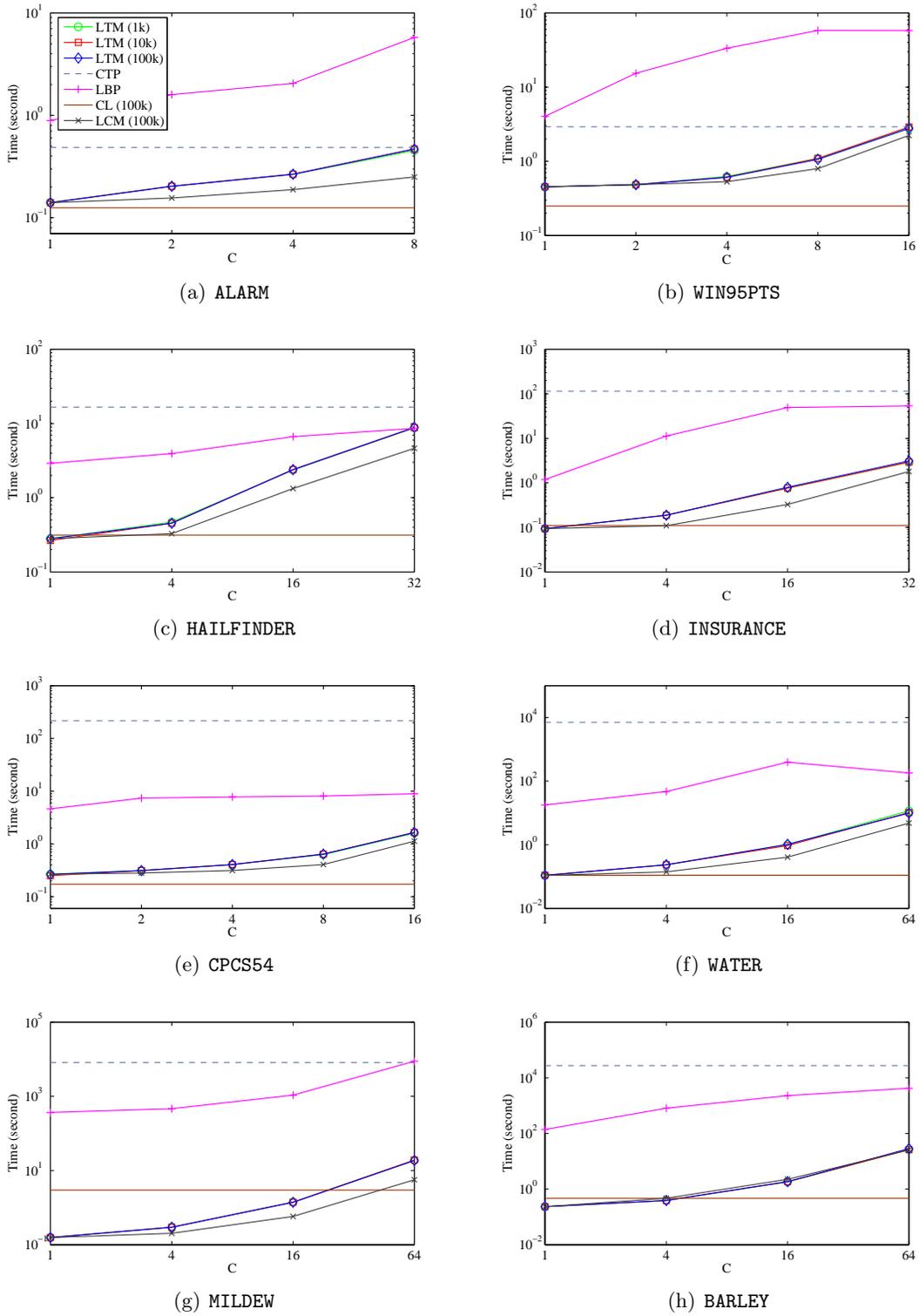

Figure 5: Running time of the online inference.

## 5.4 Comparison with CL-based Method

In this subsection, we compare our method with the CL-based method. More specifically, for each network, we learn a tree model from the $100k$ samples using the maximum spanning tree algorithm developed by Chow and Liu (1968). We then use the learned tree model to answer the queries.

The approximation accuracy of the CL-based method are shown as solid horizontal lines in the plots in Figure 4. Comparing with the CL-based method, our method achieves higher accuracy in all the networks except for MILDEW. For INSURANCE, WATER, and BARLEY, the differences are significant. For MILDEW, our method is competitive with the CL-based method. In the meantime, we notice that the CL-based method achieves good approximations in all the networks except for BARLEY. The average KL divergence is around or less than $10^{-2}$.

An obvious advantage of CL-based method is its high efficiency. This can be seen from the plots in Figure 5. In most of the plots, the CL line locates below the second data point on the LTM curve. The exception is MILDEW, for which the running time of the CL-based method is as long as our method with $C = 16$.

In summary, the results suggest that the CL-based method is a good choice for approximate inference if the online inference time is very limited. Otherwise, our method is more attractive because it is able to produce more accurate results when more time is allowed.

## 5.5 Comparison with LCM-based Method

Lowd and Domingos (2005) have previously investigated the use of LCM for density estimation. Given a data set, they determine the cardinality of the latent variable using hold-out validation, and optimize the parameters using EM. It is shown that the learned LCM achieves good model fit on a separate testing set. The LCM was also used to answer simulated probabilistic queries and the results turn out to be good.

Inspired by their work, we also learned a set of LCMs from the $100k$ samples and compared them with LTMs on the approximate inference task. Our learning strategy is slightly different. Since LCM is a special case of LTM, its inferential complexity can also be controlled by changing the cardinality of the latent variable. In our experiments, we set the cardinality such that the sum of the clique sizes in the clique tree of the LCM is roughly the same as that for the LTM learned with a chosen $C$. In this way, the inferential complexity of the two models are comparable. This can be verified by examining the LCM curves in Figure 5. We then optimize the parameters of the LCM using EM with the same setting as in the case of LTM.

As shown in Figure 4, for ALARM, WIN95PTS, CPCS54, WATER, and BARLEY, the LCM curves are located above the LTM curves. That is, our method consistently outperforms the LCM-based method for all $C$. For `HAILFINDER` and `MILDEW`, our method is worse than the LCM-based method when $C$ is small. But when $C$ becomes large, our method begins to win. For `INSURANCE`, the performance of the two methods are very close. The results suggest that unrestricted LTM is more suitable for approximation inference than LCM does.

## 6. Related Work

The idea of approximating complex BNs by simple models and using the latter to make inference has been investigated previously. The existing work mainly falls into two categories. The work in the first category approximates the joint distributions of the BNs and uses the approximation to answer all probabilistic queries. In contrast, the work in the second category is query-specific. It assumes the evidence is known and directly approximates the posterior distribution of the querying nodes.

Our method falls in the first category. We investigate the use of LTMs under this framework. This possibility has also been studied by Pearl (1988) and Sarkar (1995). Pearl (1988) develops an algorithm for constructing an LTM that is marginally equivalent to a joint distribution $P(\mathbf{X})$, assuming such an LTM exists. Sarkar (1995) studies how to build good LTMs when only approximations are amenable. Their methods, however, can only deal with the cases of binary variables.

Researchers have also explored the use of other models. Chow and Liu (1968) consider tree-structured BNs without latent variables. They develop a maximum spanning tree algorithm to efficiently construct the tree model that is closest to the original BN in terms of KL divergence. Lowd and Domingos (2005) learn an LCM to summarize a data set. The cardinality of the latent variable is determined so that the logscore on a hold-out set is maximized. They show that the learned model achieves good model fit on a separate testing set, and can provide accurate answers to simulated probabilistic queries. In both work, the approximation quality and the inferential complexity of the learned model are fixed. Our method, on the other hand, provides a parameter $C$ to let users make the tradeoff between approximation quality and inferential complexity.

The work in the second category is mainly carried out under the variational framework. The mean field method (Saul, Jaakkola, & Jordan, 1996) assumes that the querying nodes are mutually independent. It constructs an independent model that is close to the posterior distribution. As an improvement to the mean field method, the structured mean field method (Saul & Jordan, 1996) preserves a tractable substructure among the querying nodes, rather than neglecting all interactions. Bishop et al. (1997) consider another improvement, i.e., mixtures of mean field distributions. It essentially fits an LCM to the posterior distribution. All these methods directly approximate posterior distributions. Therefore, they might be more accurate than our method when used to make inference. However, these methods are evidence-specific and construct approximations online. Moreover, they involve an iterative process for optimizing the variational parameters. Consequently, the online running time is unpredictable. With our method, in contrast, one can determine the inferential complexity beforehand.

## 7. Concluding Remarks

We propose a novel scheme for BN approximate inference using LTMs. With our scheme one can trade off between the approximation accuracy and the inferential complexity. Our scheme achieves good accuracy at low costs in all the networks that we examined. In particular, it consistently outperforms LBP. We also show that LTMs are superior to LCMs when used for approximate inference.

The current bottleneck of the offline phase is parameter learning. We used EM algorithm to optimize parameters, which is known to be time consuming. The problem is especially severe when the parameter $C$ and the sample size are large. One way to speed up parameter learning is to adapt the agglomerative clustering technique for learning the cardinality of a latent variable from data (Elidan & Friedman, 2001). The basic idea is to complete the training data by setting the cardinality of the latent variable large enough and assigning each record to a latent state. In each step, one selects two states of the latent variable to merge. The process repeats until the (penalized) likelihood ceases to improve. For our parameter learning problem, we can terminate the process when the desired cardinality $C$ is achieved. We also need to deal with multiple latent variables. Since the data set is completed, we expect this method to yield a good starting point for EM in a very short time, which will in turn drastically shorten the offline phase.

## Acknowledgments

We thank Haipeng Guo and Yiping Ke for insightful discussions. We are also grateful to the anonymous reviewers for their valuable comments and suggestions on the earlier version of this paper. Research on this work was supported by Hong Kong Grants Council Grants #622105 and #622307, and the National Basic Research Program of China (aka the 973 Program) under project No. 2003CB517106. The work was completed when the first author was on leave at the HKUST Fok Ying Tung Graduate School, Guangzhou, China.